\documentclass[letterpaper]{article} 
\usepackage[draft]{aaai25}
\usepackage{amsmath}
\usepackage{amssymb}
\usepackage{bbold} 
\usepackage{times}  
\usepackage{helvet}  
\usepackage{courier}  
\usepackage[hyphens]{url}  
\usepackage{graphicx} 
\usepackage{natbib}  
\usepackage{caption} 
\usepackage{multirow}
\usepackage{lmodern,babel,adjustbox,booktabs,multirow}
\usepackage{algorithm}
\usepackage{algorithmic}
\usepackage{todonotes}
\usepackage{comment}
\usepackage{color}

\usepackage{newfloat}
\usepackage{listings}
\DeclareCaptionStyle{ruled}{labelfont=normalfont,labelsep=colon,strut=off} 
\floatstyle{ruled}
\newfloat{listing}{tb}{lst}{}
\floatname{listing}{Listing}
\title{KODA: A Data-Driven Recursive Model for Forecasting and Data Assimilation Using Koopman operator}

\title{KODA: A Data-Driven Recursive Model for Time Series Forecasting and data assimilation using koopman operators}
\author {
    Ashutosh Singh\textsuperscript{\rm 1},
    Ashish Singh\textsuperscript{\rm 3},
    Tales Imbiriba\textsuperscript{\rm 1, \rm 2},
    Deniz Erdogmus\textsuperscript{\rm 1}, 
    Ricardo Borsoi\textsuperscript{\rm 4}
}
\affiliations {
    \textsuperscript{\rm 1}Department of Electrical and Computer Engineering, Northeastern University, USA\\
    \textsuperscript{\rm 2}Institute for Experiential AI, Northeastern University, USA\\
    \textsuperscript{\rm 3}CICS, University of Massachusetts Amherst, USA 
    \textsuperscript{\rm 4}CNRS, CRAN, Université de Lorraine, France\\
    \texttt{\{singh.ashu,d.erdogmus,talesim\}@northeastern.edu}, \texttt{ashishsingh@cs.umass.edu}, \texttt{raborsoi@gmail.com}
}

\begin{document}

\maketitle

\begin{abstract}
Approaches based on Koopman operators have shown great promise in forecasting time series data generated by complex nonlinear dynamical systems (NLDS). Although such approaches are able to capture the latent state representation of a NLDS, they still face difficulty in long term forecasting when applied to real world data. Specifically many real-world NLDS exhibit time-varying behavior, leading to nonstationarity that is hard to capture with such models. Furthermore they lack a systematic data-driven approach to perform data assimilation, that is, exploiting noisy measurements on the fly in the forecasting task. To alleviate the above issues, we propose a Koopman operator-based approach (named KODA - \textbf{K}oopman \textbf{O}perator with \textbf{D}ata \textbf{A}ssimilation) that integrates forecasting and data assimilation in NLDS. In particular we use a Fourier domain filter to disentangle the data into a physical component whose dynamics can be accurately represented by a Koopman operator, and residual dynamics that represents the local or time varying behavior that are captured by a flexible and learnable recursive model. We carefully design an architecture and training criterion that ensures this decomposition lead to stable and long-term forecasts. Moreover, we introduce a course correction strategy to perform data assimilation with new measurements at inference time. The proposed approach is completely data-driven and can be learned end-to-end. Through extensive experimental comparisons we show that KODA outperforms existing state of the art methods on multiple time series benchmarks such as electricity, temperature, weather, lorenz 63 and duffing oscillator demonstrating its superior performance and efficacy along the three tasks a) forecasting, b) data assimilation and c) state prediction.
\end{abstract}

%

\section{Introduction}
Many real world applications tackle phenomena which are dynamic in nature. Measuring and tracking the evolution of such phenomena holds extreme importance in many fields such as weather forecasting \cite{pathak2022fourcastnet,lam2022graphcast}, evolution of PDEs \cite{li2020fourier,kovachki2023neural}, modeling neural dynamics \cite{brunton2016extracting,singh2021variation}, etc. Most of these phenomena are characterised as nonlinear dynamical systems (NLDS). If prior knowledge about the underlying system's evolution is known then it can inform modeling, see, e.g., \cite{cai2021physics,cuomo2022scientific,raissi2019physics,imbiriba2023augmented}. However, in most scenarios no direct information is available about the physics of the underlying system. Therefore, for such cases we can only depend on the measurement data to model system's dynamics \cite{guen2020disentangling,afshar2023extended,frion2024koopman}. Key motivations for such modeling is to evolve the model and generate accurate forecasts over long horizon. But the non-linear and non-stationary characteristic of these system makes it hard to capture the state representation or generate accurate forecasts. One classical approach is to model the NLDS in a representation space where the system dynamics is linear. Using the tools from spectral theory and linear algebra we can study various characteristics of the NLDS such as stability, asymptotic nature etc. 

\begin{figure}
    \centering
    \includegraphics[width=\linewidth]{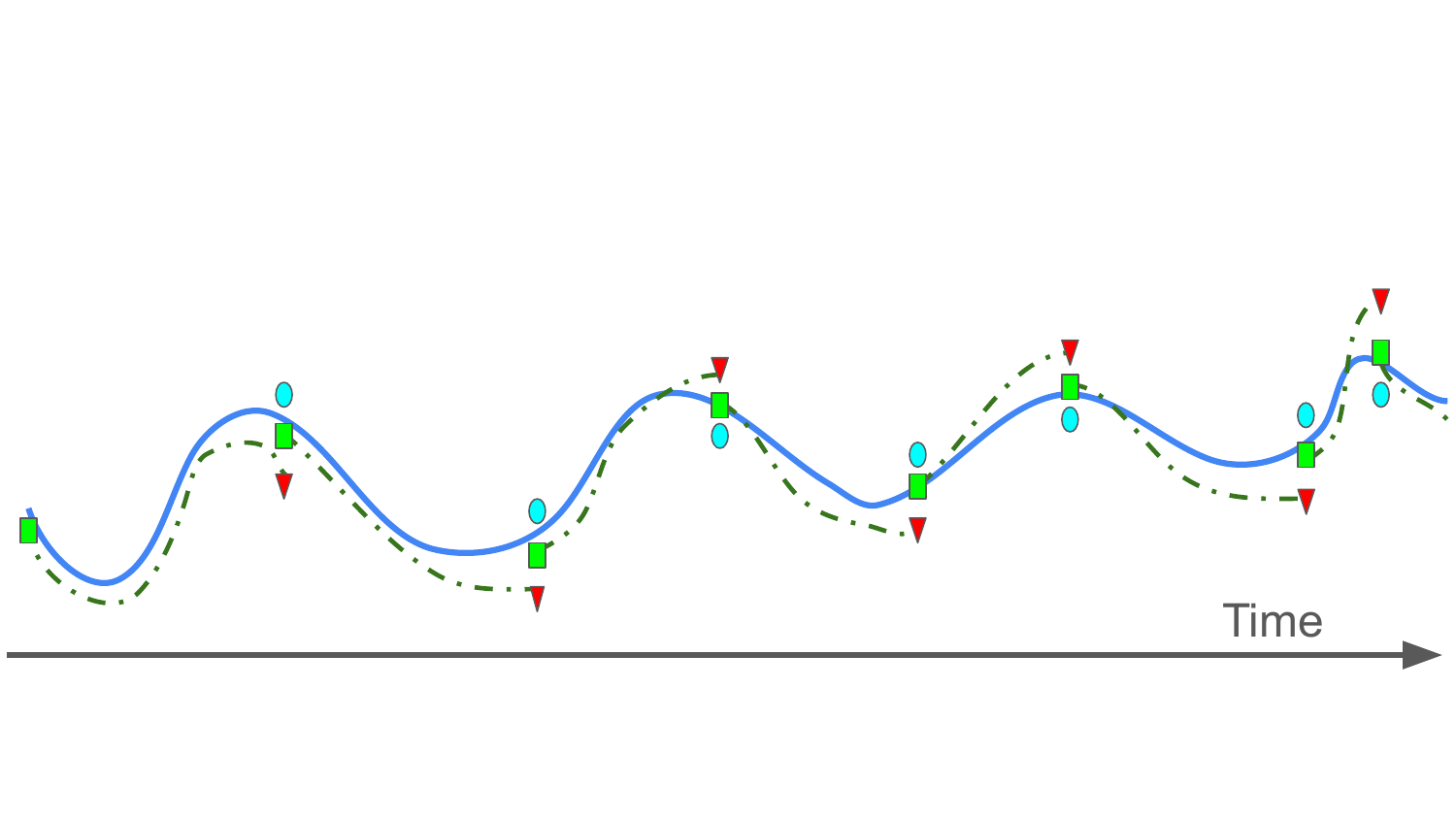}
    \vspace{-8ex}
    \caption{Pictorial representation of data assimilation using a recursive prediction-correction. The blue line represents actual state trajectory while the dotted line represents the trajectory generated by the learned model. Red triangles represent the model prediction while the green squares represent the corrected prediction using the noisy measurement represented by the blue circles. }
    \label{fig:DA}
\end{figure}

 Koopman operator theory \cite{koopman1931hamiltonian,phillips1961generalized} generates a representation with a linear dynamics from the NLDS. It achieves this by mapping the system’s dynamics into an infinite-dimensional space using a set of measurement functions. Given that learning such a mapping is computationally infeasible, numerous algorithms have been proposed in the literature which seek to approximate the Koopman operator in finite dimensions and effectively describe the NLDS \cite{brunton2021modern}. In recent years, deep learning algorithms have emerged to learn these operators directly from the measurement snapshot data to predict the system’s state space evolution. Despite these advancements, such methods are still susceptible to error accumulation, which can lead to significant drift in state space estimation during long-term forecasting. This calls for an adaptive framework that can utilise measurement data on the fly to correct its prediction. \ref{fig:DA} represents a prediction-correction mechanism for modeling a dynamical system. When measurements snapshots are available they are used to correct prediction error.  

The Kalman filter presents a systematic way for state space prediction and correction mechanism using measurement data. Many deep learning methods have been recently proposed that takes inspiration from the traditional Kalman filter framework, motivating the use of neural network architectures in an adaptive manner to produce robust long term forecasts \cite{revach2022kalmannet,singh2024learning,guen2020disentangling}.

In this paper, we propose a branched architecture that disentangles the dynamics into global and local (residual) components in a segmented fashion. This allows us to recover the dynamics of the underlying global factor while being robust to noise and local variations within each segment. Additionally, by not discarding the local sources of variation and capturing it with a recurrent residual model we are able to generate long term time series forecasting (LTSF). We further motivate a data assimilation framework for this disentangled view of the state space to perform online data assimilation during inference.
The main contributions of this work are:
\begin{itemize}
    \item The proposed approach is able to disentangle the physical and residual components of the dynamics and evolve them separately to generate forecasts.
    \item We introduce a Kalman filter-inspired data assimilation framework with the Koopman operator-based prediction model.
    \item The proposed approach can leverage data available during inference time to improve the prediction of the model hence making it possible to generate accurate forecasts over long horizons, hence presenting a solution for LTSF.
    \item The proposed approach is one of the first works that extends data assimilation to a disentangled state representation-based model.
\end{itemize}

\section{Related Works}
\subsection{Learning with Koopman Operator}
Koopman operator theory \cite{brunton2021modern} serves as a potent tool for unveiling the inherent dynamics of non-linear systems. Notably, the recent surge in interest surrounding Koopman operators can be attributed to the strong theoretical foundations and empirical success of such algorithms. Most famous among them is Dynamic Mode Decomposition \cite{tu2013dynamic}, a data-driven approach \cite{brunton2016discovering,kutz2016dynamic} enabling a practical approximation of the Koopman operator.
Recent works (see, e.g., \cite{kawahara2016dynamic,lusch2018deep,yeung2019learning,takeishi2017learning,bevanda2023koopman}) further explored learning theory for estimating Koopman operators. In particularly, the Koopman autoencoder architecture proposed in \cite{azencot2020forecasting} and \cite{lusch2018deep} has become one of the most widely used deep learning models for this task. Since then several work have been proposed in the intersection of learning theory and Koopman operators \cite{fathi2024course,Wang_2023_CVPR,berman2023multifactor}.
The recent models proposed in \cite{liu2024koopa,wang2023koopman}, proposed Koopman-based neural forecasting methods by disentangling the system's dynamics into local and global components. However, they lack a systematic approach for data assimilation during inference time. 

\subsection{Data Assimilation}
Some of the earlier work that combine Koopman operators with the Kalman filter are \cite{benosman2017koopman}, where the author formulates a linear observer design using Koopman operators to predict crowd flow, and \cite{jiang2022data} where the authors use a kernel-based Koopman operator for robotic systems. Unlike these works, the proposed approach is a deep learning-based recursive approach. \cite{frion2024neural} is one of the first works that motivates data assimilation with a neural Koopman operator. While the model proposed in \cite{frion2024neural} uses a Koopman operator as prior for variational data assimilation, in KODA we adopt a branched prediction model within a Kalman filter-inspired framework for both forecasting and assimilation. Additionally, unlike \cite{frion2024neural}, KODA can achieve online data assimilation during inference. Similarly, in \cite{singh2024learning} the authors propose a Kalman filter-based data assimilation using neural operators for semilinear PDEs. Differently, our method motivates data assimilation jointly with a Koopman operator-based prediction model and a parallel residual model. We also don't assume access to a ground truth of a latent state representation of the system and instead learn to evolve the model just from measurement data. In \cite{frion2024koopman} authors propose a way of uncertainty quantification for time series forecasting using Koopman ensembles.

\section{Problem Setting}
In this section, we formally define the problem of learning a model that can not only produce solution for LTSF problem, provided sufficient historical data, but also utilise new data available during inference to correct the model prediction. The underlying physical quantity whose dynamic we observe is referred to as states and the observed data itself is referred to as measurements of the state. 
Therefore, let us denote $s(t) \in \mathcal{H}$ as the state representation, and by $y(t) \in \mathcal{Y}$ as the representation of the measurement snapshots. $\mathcal{H}$ and $\mathcal{Y}$ are finite dimensional vector spaces.
Hence a NLDS can be described as:
\begin{align}
    \frac{\partial s(t)}{\partial t} = \mathcal{A}(s(t)) + \eta(t)
    \\
    y(t) = \mathcal{B}(s(t)) + \epsilon(t)
    \label{eq:sys}
\end{align}
where operator $\mathcal{A}:\mathcal{H}\to\mathcal{H}$ models the evolution of the states, $\mathcal{B}:\mathcal{H}\to\mathcal{Y}$ represents the mapping from the states to the measurements, $\eta(t)\in\mathcal{H}$ denote a process noise which represents possible stochasticity in the state evolution, and $\epsilon(t)\in\mathcal{Y}$ denotes measurement noise.\\
The main objective of the forecasting problem is to obtain accurate estimates of the future measurement trajectory $\tilde{y}_{t+\tau}=\mathcal{B}(s(t+\tau))$, $\tau\in\{1,2,3,\ldots\}$ based on past measurements $\{y_0,\ldots,y_t\}$, where $y_t$ denote a discrete measurement of $y(t)$. Note that while the NLDS~\eqref{eq:sys} might be described in continuous time, we consider the discrete-time forecasting problem, based on discrete measurements. In order to keep notation clear, we denote the discrete time index as a subscript. Considering some training dataset with different realizations of $\{y_1,\ldots,y_T\}$, the forecasting objective can be formulated as learning an operator $\Phi:\mathcal{H}^t\times\mathbb{N}_*\to\mathcal{H}$ that predicts the future measurements as $\tilde{y}_{t+\tau}=\Phi(y_1,\ldots,y_t;\tau)$. The above objective is generally optimised by minimizing the prediction loss between the forecast and ground truth during training. The data assimilation problem extends this task by including the possibility of incorporating a sparse set of measurements, irregularly present along the forecast horizon, on the fly to improve the forecast estimate.

Note that learning such a model is not easy since the amount of training data might be limited, and the NLDS under consideration can be highly nonlinear. Moreover the measurements $y(t)$ are often accompanied with noise. This makes it harder to avoid drift in the forecast over time. 
In the section below, we introduce KODA a framework for generating long term forecasts and preforming data assimilation, while enabling us to study the properties of the NLDS through the learned koopman operator.

\section{KODA}
\begin{figure}
    \centering
    \includegraphics[width=\linewidth]{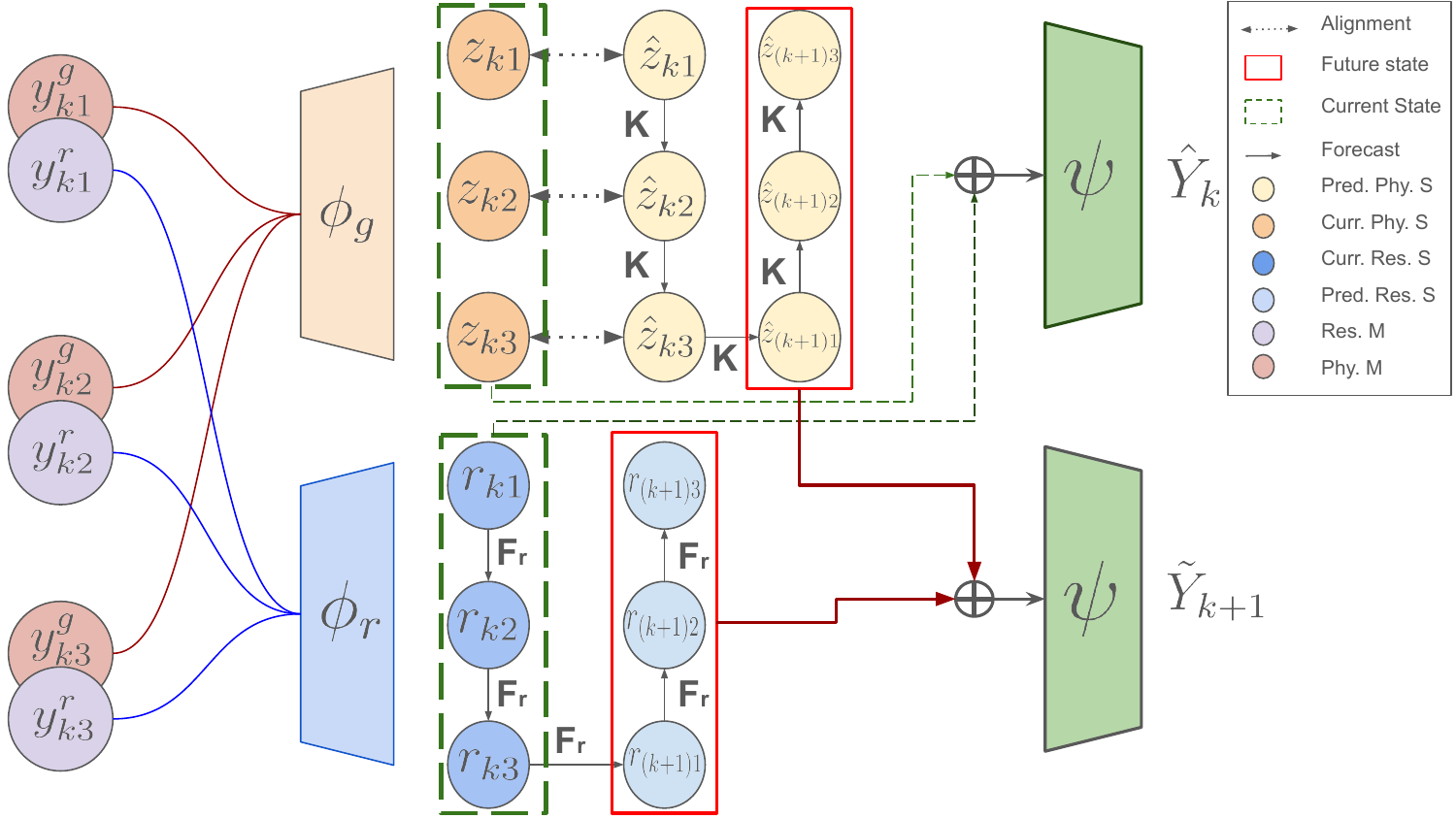}
    
    \caption{Schematic diagram of one time step prediction from KDOA. Z represent physical state component, R represents the residual while Y represents measurement data. $F_r$ and $K$ rperesents prediction models.}
    \label{fig:sys_daig}
\end{figure}

In this section, we formally propose KODA and its various elements. KODA adopts a disentangled view of the state representation separating it into a physical and a residual component \ref{fig:sys_daig}. To achieve this purely from the measurement data, we separate the dominant spectrum across the data using a Fourier filter. We then use an encoding model that learns the physical and residual components, of the state representation, from the dominant and non-dominant parts of the measurement data. Using separate prediction models, for both the state space components, we generate their future estimates. Upon adding the predicate from the two models we compute the future state estimates. We then use a decoder to map the state estimate back to the measurement space, hence recovering the future forecast. We perform this recursively in a windowed fashion to generate future trajectory. At inference time whenever the measurement data is available we use it in the correction mechanism of KODA to update the model predictions. Hence KODA presents a framework that is purely learning based and data driven.  

\subsection{Data Encoding Scheme}
\subsubsection{State Disentanglement}
In order to facilitate the task of LTSF, we aim to learn a latent representation $h_t\in\mathcal{H}$. This representation act as the learning based approximation of the true $s_t$ in~\eqref{eq:sys}. Assume that $h_t$ comprises of multiple sources of variation (physical and residual) we use the following disentanglement of the state trajectory,
\begin{align}
    h_t = z_t + r_t \,,
    \label{eq:state_disentangle}
\end{align}
where $z_t \in \mathcal{H}$ represents the underlying physical component while $r_t \in \mathcal{H}$ represents the residual element of the state trajectory. By disentangling the dynamics into these two components we aim at recovering the physical component i.e. the stable component (time invariant in the selected window). The benefit of such disentanglement is that we are able to separate physical component from the local ones. The physical component is expected to capture sources of variation that cause long term change and are stable in smaller window analysis of which could shed light on the spectral property of the underlying system. Due to its usefulness this disentangled view of the dynamics have garnered interest in recent literature \cite{guen2020disentangling}, \cite{liu2024koopa} and \cite{wang2023koopman}. 

\subsubsection{Measurement Disentanglement}
Since no ground truth is available about the true state, we learn to approximate the state representations purely from the measurement data. We use Fourier filtering to find and separate the dominant spectrum present across the measurement space. To achieve the best performance for LTSF we pre-computed a filter by taking all the $\tau$ length windows in the training data and extracted the dominant spectrum across them. We define $Y_k = \{y_t,y_{t+1}, ..., y_{t+\tau}\}$ as the sliding window over the data with $\tau\in\mathbb{N}_*$ as the window length. We use this Fourier filtering to represent the resulting disentanglement in the measurement space as, 
\begin{align}
  Y^g_k &= \mathcal{F}^{-1}(G_{\alpha}(\mathcal{F}(Y_k)))\\
  Y^r_k &= \mathcal{F}^{-1}(\overline{G}_{\alpha}(\mathcal{F}(Y_k)))\\
  \text{s.t.} \quad  Y_k &= Y^g_k + Y^r_k
  \label{eq:measurement_disentanglement}
\end{align}
here $G_{\alpha}$ represents the filter, where $\alpha$ is a hyper-parameter, and $\overline{G}_{\alpha}$ represents its complement which is designed such that~\eqref{eq:measurement_disentanglement} is satisfied, guaranteeing no information loss. $\mathcal{F}$ and $\mathcal{F}^{-1}$ represents Fourier transform and its inverse applied in the time dimension. Both $y^g_t$ and $y^r_t$ are defined such that  
$y^g_t \in \mathcal{Y}$ and $y^r_t \in \mathcal{Y}$. We assume that whenever the data is available it always exist as a block of $\tau$ samples. In many application, such as electricity, motor fault analysis etc, it is quite common that the data is collected for a small window of time at irregular intervals. Hence during the data assimilation task we assume that data is irregularly present as a block of $\tau$.

\subsubsection{Segmentation and Encoding}
Within each $k^{th}$ window we create further $s$ segments of segment length $w =\tau/s$. Hence $Y^g_k = \{\mathbb{y}^g_{k1}, \mathbb{y}^g_{k2}, ..., \mathbb{y}^g_{ks}\}$ after reshaping, where $\mathbb{y}^g_{ki}$ represents the $i^{th}$ segment of $k^{th}$ window. Same segmentation is used for $Y^r_k$. Therefore both $\mathbb{y}^g_{ki} \in \mathcal{Y}^{w}$ and $\mathbb{y}^r_{ki} \in \mathcal{Y}^{w}$.  
We use identical encoder architecture $\phi_g: \mathcal{Y}^{w}\rightarrow \mathcal{H}$ and $\phi_r: \mathcal{Y}^{w} \rightarrow \mathcal{H}$ for $Y^g_k$ and $Y^r_k$ respectively, to then get $Z_k = \{z_{k1},z_{k2}, ..., z_{ks}\}$ and $R_k = \{r_{k1},r_{k2}, ..., r_{ks}\}$ the respective representations as
\begin{align}
    z_{ki} = \phi_g(\mathbb{y}^g_{ki})  \,, \quad r_{ki} = \phi_r(\mathbb{y}^r_{ki})  \,.
    \label{eq:encoding}
\end{align}
 Previous work \cite{lin2023segrnn} have shown that segment wise iteration leads to better forecast for the LTSF problem instead of point wise iteration. 

\subsection{Prediction Model}
Based on the disentanglement of the latent state representation \eqref{eq:state_disentangle}, we design a prediction model which is used to propagate the states \eqref{eq:encoding} forward along the temporal dimension. The prediction model consists of two parallel branches: (i) a Global model, which propagates the component $Z_k$, and (ii) a Residual model which propagates the component $R_k$. We describe each of these models below. 
\subsubsection{Residual Model}
Modeling the residual is important for the reconstruction and for the forecast of the measurement trajectory. We can define the residual $R_t$ trajectory as follows,
\begin{equation}
    r_{ki} = F_r(r_{k(i-1)})
    \label{eq:residualpred}
\end{equation}
here $F_r : \mathcal{H}\rightarrow\mathcal{H}$ represents the residual model. We use a gated recurrent unit (GRU) \cite{chung2014empirical} to then propagate the residual trajectory. Other recurrent architecture such as LSTM \cite{cheng2016long} or RNN \cite{salehinejad2017recent} could also be used here.
\subsubsection{Physical Model}
 Our prediction model for the physical component of the signal learns koopman operator $\mathcal{K}:\mathcal{H} \rightarrow \mathcal{H}$ such that,
\begin{align}
    \hat{z}_{k(i)} &= \mathcal{K}\circ \hat{z}_{k(i-1)}
    \label{eq:globalpred}
\end{align}
many methods in literature has been proposed to estimate the Koopman operator from data. Note that $\hat{z}_{ki}$ in \eqref{eq:globalpred} and $z_{ki}$ in \eqref{eq:encoding} are different as the later represents the encoding from the measurement while the prior represents the koopman forecast. To align the two representation, an alignment loss is computed and minimised. In \cite{lusch2018deep,azencot2020forecasting} the $\mathcal{K}$ is composed recursively with $\hat{z}_0$ to generate all the future $\hat{z}$. This leads to poor performance and drift especially for multivariate high dimensional non-stationary data as shown in \cite{fathi2024course}. This also leads to poor generalisation across state space not observed in the look-back window. To mitigate this, during inference for LTSF task we make use of periodic re-encoding that is after every $k^{th}$ windows we use the decoder to generate predicted measurement window $(k+1)^{th}$ and use the encoding model to get state representation.

Upon adding the output of \eqref{eq:globalpred} and \eqref{eq:residualpred} we can get the state representation for the next segment. For each prediction cycle we use \eqref{eq:globalpred} and \eqref{eq:residualpred} to predict all the segments in the future window and stack them in order to get $\hat{Z}_k$ and $R_k$ respectively. Similarly we stack and add the output of the encoders in \eqref{eq:encoding} to get Reconstruction of $(k-1)^{th}$ window.
\begin{equation}
    \tilde{Y}_{k} = \psi(\hat{Z}_k + R_k), \quad \hat{Y}_{k-1} = \psi(Z_{k-1} + R_{k-1})
    \label{eq:jointpred}
\end{equation}
here $\psi:\mathcal{H}^{s} \rightarrow  \mathcal{Y}^{s\times w}$ is the decoder that maps from state space to the segmented view of the measurement space. We reshape $\tilde{Y}_k$ and $\hat{Y}_{k-1}$ back to $\mathcal{Y}^{\tau}$.
 Therefore by continuously evolving the physical and residual components and decoding their output \eqref{eq:jointpred} we can  generate $\tilde{Y}_k$ and reconstructed trajectory $\hat{Y}_{k-1}$.

\subsection{Correction Model}
\begin{table*}[!htbp]
\centering
\caption{Multivariate long-term time series forecasting results. We evaluate all the methods across forecast horizons of \(H \in \{96, 192, 336, 720\}\) in all datasets. We take the average of 10 iterations for KODA. \textbf{Bold} represents the best and \underline{underlined} represents the second best.} 
\begin{adjustbox}{max width=\textwidth}
\begin{tabular}{cc|llcc|cccccc|cccc|llll|cccc}
\hline
\multicolumn{2}{c|}{Models}                                                                & \multicolumn{2}{c}{\begin{tabular}[c]{@{}c@{}}KODA\\ (ours)\end{tabular}} & \multicolumn{2}{c|}{\begin{tabular}[c]{@{}c@{}}SegRNN\\ (2023)\end{tabular}} & \multicolumn{2}{c}{\begin{tabular}[c]{@{}c@{}}PatchTST\\ (2023)\end{tabular}} & \multicolumn{2}{c}{\begin{tabular}[c]{@{}c@{}}FEDformer\\ (2022)\end{tabular}} & \multicolumn{2}{c|}{\begin{tabular}[c]{@{}c@{}}Informer\\ (2021)\end{tabular}} & \multicolumn{2}{c}{\begin{tabular}[c]{@{}c@{}}TiDE\\ (2023)\end{tabular}} & \multicolumn{2}{c|}{\begin{tabular}[c]{@{}c@{}}Dlinear\\ (2023)\end{tabular}} & \multicolumn{2}{c}{\begin{tabular}[c]{@{}c@{}}KOOPA\\ (2023)\end{tabular}} & \multicolumn{2}{c|}{\begin{tabular}[c]{@{}c@{}}KNF\\ (2022)\end{tabular}} & \multicolumn{2}{c}{\begin{tabular}[c]{@{}c@{}}MICN\\ (2023)\end{tabular}} & \multicolumn{2}{c}{\begin{tabular}[c]{@{}c@{}}TimesNet\\ (2023)\end{tabular}} \\ \hline
\multicolumn{2}{c|}{Metric}                                                                & \multicolumn{1}{c}{MSE}             & \multicolumn{1}{c}{MAE}             & MSE                                   & MAE                                  & MSE                                  & MAE                                    & MSE                                    & MAE                                   & MSE                                    & MAE                                   & MSE                                 & MAE                                 & MSE                                   & MAE                                   & MSE                         & \multicolumn{1}{c}{MAE}                      & \multicolumn{1}{c}{MSE}             & \multicolumn{1}{c|}{MAE}            & MSE                                 & MAE                                 & MSE                                   & MAE                                   \\ \hline
\multicolumn{1}{c|}{\multirow{4}{*}{\rotatebox{90}{ETTh1}}}       & 96  & \textbf{0.339}                      & \textbf{0.359}                      &  0.341                           &  0.376                         & 0.376                                & 0.402                                  & 0.376                                  & 0.415                                 & 0.941                                  & 0.769                                 & 0.375                               & 0.398                               & 0.374                                 & 0.399                                 & 0.374                       & 0.408                                        & 0.457                               & 0.459                               & 0.421                               & 0.431                               & 0.384                                 & 0.402                                 \\
\multicolumn{1}{c|}{}                                                                & 192 & \textbf{0.356}                      & \textbf{0.388}                      &  0.385                          &  0.402                          & 0.413                                & 0.429                                  & 0.423                                  & 0.446                                 & 1.007                                  & 0.786                                 & 0.412                               & 0.422                               & 0.405                                 & 0.416                                 & 0.425                       & 0.446                                        & 0.518                               & 0.520                               & 0.474                               & 0.487                               & 0.436                                 & 0.429                                 \\
\multicolumn{1}{c|}{}                                                                & 336 & \textbf{0.394}                      & \textbf{0.404}                      &  0.403                          &  0.417                          & 0.429                                & 0.438                                  & 0.444                                  & 0.462                                 & 1.038                                  & 0.784                                 & 0.435                               & 0.433                               & 0.439                                 & 0.443                                 & 0.437                       & 0.472                                        & 0.592                               & 0.551                               & 0.569                               & 0.551                               & 0.491                                 & 0.469                                 \\
\multicolumn{1}{c|}{}                                                                & 720 & \textbf{0.411}                      & 0.451                         & 0.434                           & \textbf{0.447}                       & 0.448                                & 0.468                                  & 0.469                                  & 0.492                                 & 1.144                                  & 0.857                                 & 0.454                               & 0.465                               & 0.472                                 & 0.49                                  & 0.461                       & 0.498                                        & 0.784                               & 0.596                               & 0.775                               & 0.672                               & 0.521                                 & 0.538                                 \\ \hline
\multicolumn{1}{c|}{\multirow{4}{*}{\rotatebox{90}{ETTh2}}}       & 96  & \textbf{0.255}                      & 0.324                       & 0.263                           & \textbf{0.321}                       & 0.274                                & 0.337                                 & 0.332                                  & 0.374                                 & 1.549                                  & 0.952                                 & 0.284                               & 0.336                               & 0.289                                 & 0.353                                 & 0.297                       & 0.355                                        & 0.433                               & 0.446                               & 0.299                               & 0.364                               & 0.34                                  & 0.374                                 \\
\multicolumn{1}{c|}{}                                                                & 192 & \textbf{0.316}                      & \textbf{0.361}                      & 0.321                        & 0.367                       & 0.341                                & 0.382                                  & 0.407                                  & 0.446                                 & 3.792                                  & 1.542                                 & 0.332                               & 0.381                               & 0.383                                 & 0.418                                 & 0.356                       & 0.394                                        & 0.528                               & 0.503                               & 0.441                               & 0.454                               & 0.402                                 & 0.414                                 \\
\multicolumn{1}{c|}{}                                                                & 336 &0.327                         & 0.385                               & \textbf{0.325}                        & \textbf{0.374}                       & 0.329                                & 0.384                            & 0.410                                  & 0.447                                 & 4.215                                  & 1.642                                 & 0.365                               & 0.407                               & 0.448                                 & 0.465                                 & 0.385                       & 0.421                                        & 0.631                               & 0.617                               & 0.654                               & 0.567                               & 0.452                                 & 0.452                                 \\
\multicolumn{1}{c|}{}                                                                & 720 & \textbf{0.362}                      & \textbf{0.416}                      & 0.394                                 & 0.424                                & 0.379                          & 0.422                           & 0.412                                  & 0.469                                 & 3.656                                  & 1.619                                 & 0.419                               & 0.451                               & 0.605                                 & 0.551                                 & 0.427                       & 0.476                                        & 0.795                               & 0.738                               & 0.956                               & 0.716                               & 0.462                                 & 0.468                                 \\ \hline
\multicolumn{1}{c|}{\multirow{4}{*}{\rotatebox{90}{ETTm1}}}       & 96  & \textbf{0.281}                      & \textbf{0.331}                      & 0.284                           & 0.339                                & 0.293                                & 0.336                           & 0.326                                  & 0.39                                  & 0.626                                  & 0.56                                  & 0.306                               & 0.349                               & 0.299                                 & 0.344                                 & 0.296                       & 0.344                                        & 0.341                               & 0.394                               & 0.316                               & 0.362                               & 0.338                                 & 0.375                                 \\
\multicolumn{1}{c|}{}                                                                & 192 & 0.323                        & \textbf{0.362}                      & \textbf{0.319}                        & 0.375                                & .335                                 & 0.371                                  & 0.365                                  & 0.415                                 & 0.725                                  & 0.619                                 & 0.335                               & 0.366                        & 0.335                                 & 0.368                                 & 0.339                       & 0.378                                        & 0.414                               & 0.428                               & 0.363                               & 0.390                               & 0.374                                 & 0.387                                 \\
\multicolumn{1}{c|}{}                                                                & 336 & \textbf{0.341}                      & \textbf{0.379}                      & 0.349                          & 0.383                                & 0.369                                & 0.392                                  & 0.392                                  & 0.425                                 & 1.005                                  & 0.741                                 & 0.364                               & 0.381                         & 0.369                                 & 0.386                                 & 0.381                       & 0.392                                        & 0.426                               & 0.440                               & 0.408                               & 0.426                               & 0.410                                 & 0.411                                 \\
\multicolumn{1}{c|}{}                                                                & 720 & 0.409                       &0.420                         & \textbf{0.407}                        & \textbf{0.418}                       & 0.416                                & 0.420                                  & 0.446                                  & 0.458                                 & 1.133                                  & 0.845                                 & 0.413                               & \textbf{0.418}                      & 0.425                                 & 0.421                                 & 0.435                       & 0.438                                        & 0.499                               & 0.467                               & 0.481                               & 0.476                               & 0.478                                 & 0.458                                 \\ \hline
\multicolumn{1}{c|}{\multirow{4}{*}{\rotatebox{90}{ETTm2}}}       & 96  & \textbf{0.154}                      & \textbf{0.237}                      & 0.158                           & 0.241                       & 0.166                                & 0.256                                  & 0.182                                  & 0.271                                 & 0.355                                  & 0.462                                 & 0.161                               & 0.251                               & 0.167                                 & 0.260                                 & 0.171                       & 0.254                                        & 0.312                               & 0.328                               & 0.179                               & 0.275                               & 0.187                                 & 0.267                                 \\
\multicolumn{1}{c|}{}                                                                & 192 & \textbf{0.211}                      & \textbf{0.278}                      & 0.215                          & 0.283                                & 0.223                                & 0.296                                  & 0.252                                  & 0.318                                 & 0.595                                  & 0.586                                 & 0.215                         & 0.281                      & 0.224                                 & 0.303                                 & 0.226                       & 0.298                                        & 0.425                               & 0.490                               & 0.307                               & 0.376                               & 0.249                                 & 0.309                                 \\
\multicolumn{1}{c|}{}                                                                & 336 & 0.267                       & 0.325                         & \textbf{0.263}                        & \textbf{0.317}                       & 0.274                                & 0.329                                  & 0.324                                  & 0.364                                 & 1.277                                  & 0.871                                 & 0.267                         & 0.326                               & 0.281                                 & 0.342                                 & 0.287                       & 0.362                                        & 0.491                               & 0.547                               & 0.325                               & 0.388                               & 0.321                                 & 0.351                                 \\
\multicolumn{1}{c|}{}                                                                & 720 & \textbf{0.322}                      & \textbf{0.358}                      & 0.328                           & 0.366                          & 0.362                                & 0.385                                  & 0.410                                  & 0.424                                 & 3.001                                  & 1.267                                 & 0.352                               & 0.383                               & 0.397                                 & 0.421                                 & 0.391                       & 0.406                                        & 0.568                               & 0.612                               & 0.502                               & 0.493                               & 0.408                                 & 0.403                                 \\ \hline
\multicolumn{1}{c|}{\multirow{4}{*}{\rotatebox{90}{Electricity}}} & 96  & 0.131                               & 0.220                         & \textbf{0.128}                        & \textbf{0.219}                       & 0.129                          & 0.222                                  & 0.186                                  & 0.302                                 & 0.304                                  & 0.393                                 & 0.132                               & 0.229                               & 0.142                                 & 0.237                                 & 0.136                       & 0.236                                        & 0.198                               & 0.284                               & 0.164                               & 0.269                               & 0.168                                 & 0.272                                 \\
\multicolumn{1}{c|}{}                                                                & 192 & \textbf{0.144}                      & \textbf{0.225}                      & 0.148                                 & 0.239                          & 0.147                         & 0.249                                  & 0.197                                  & 0.311                                 & 0.327                                  & 0.417                                 & 0.147                       & 0.243                               & 0.153                                 & 0.249                                 & 0.156                       & 0.255                                        & 0.245                               & 0.320                               & 0.177                               & 0.285                               & 0.184                                 & 0.289                                 \\
\multicolumn{1}{c|}{}                                                                & 336 & \textbf{0.157}                      & \textbf{0.251}                      & 0.166                                 & 0.258                          & 0.163                                & 0.259                                  & 0.213                                  & 0.328                                 & 0.333                                  & 0.422                                 & 0.161                        & 0.261                               & 0.169                                 & 0.267                                 & 0.178                       & 0.281                                        & 0.281                               & 0.351                               & 0.193                               & 0.304                               & 0.198                                 & 0.309                                 \\
\multicolumn{1}{c|}{}                                                                & 720 & \textbf{0.189}                      & 0.293                         & 0.201                                 & 0.298                                & 0.197                                & \textbf{0.291}                         & 0.233                                  & 0.344                                 & 0.351                                  & 0.427                                 & 0.196                        & 0.294                               & 0.203                                 & 0.301                                 & 0.210                       & 0.314                                        & 0.304                               & 0.387                               & 0.212                               & 0.321                               & 0.22                                  & 0.321                                 \\ \hline
\multicolumn{1}{c|}{\multirow{4}{*}{\rotatebox{90}{Traffic}}}     & 96  & 0.397                               & \textbf{0.227}                      & 0.543                                 & 0.235                          & 0.361                         & 0.249                                  & 0.576                                  & 0.359                                 & 0.733                                  & 0.412                                 & \textbf{0.334}                      & 0.253                               & 0.410                                 & 0.282                                 & 0.401                       & 0.275                                        & 0.643                               & 0.376                               & 0.519                               & 0.309                               & 0.593                                 & 0.321                                 \\
\multicolumn{1}{c|}{}                                                                & 192 & 0.396                               & 0.248                         & 0.567                                 & \textbf{0.242}                       & 0.379                          & 0.256                                  & 0.619                                  & 0.381                                 & 0.777                                  & 0.435                                 & \textbf{0.342}                      & 0.257                               & 0.423                                 & 0.287                                 & 0.405                       & 0.284                                        & 0.699                               & 0.405                               & 0.537                               & 0.315                               & 0.617                                 & 0.336                                 \\
\multicolumn{1}{c|}{}                                                                & 336 & 0.403                               & \textbf{0.260}                      & 0.602                                 & 0.262                                & 0.393                          & 0.264                                  & 0.608                                  & 0.375                                 & 0.776                                  & 0.434                                 & \textbf{0.355}                      & 0.261                         & 0.436                                 & 0.296                                 & 0.442                       & 0.301                                        & 0.728                               & 0.429                               & 0.534                               & 0.313                               & 0.629                                 & 0.336                                 \\
\multicolumn{1}{c|}{}                                                                & 720 & 0.428                        & 0.277                         & 0.671                                 & 0.281                                & 0.432                                & 0.286                                  & 0.621                                  & 0.375                                 & 0.827                                  & 0.466                                 & \textbf{0.388}                      & \textbf{0.273}                      & 0.466                                 & 0.315                                 & 0.468                       & 0.309                                        & 0.782                               & 0.440                               & 0.577                               & 0.325                               & 0.641                                 & 0.352                                 \\ \hline
\multicolumn{1}{c|}{\multirow{4}{*}{\rotatebox{90}{Weather}}}     & 96  &  0.147                         & 0.193                        & \textbf{0.142}                        & \textbf{0.181}                       & 0.149                                & 0.198                                  & 0.238                                  & 0.314                                 & 0.354                                  & 0.405                                 & 0.166                               & 0.222                               & 0.176                                 & 0.237                                 & 0.154                       & 0.205                                        & 0.293                               & 0.308                               & 0.161                               & 0.229                               & 0.172                                 & 0.226                                 \\
\multicolumn{1}{c|}{}                                                                & 192 & \textbf{0.184}                      & 0.239                               & 0.187                          & \textbf{0.227}                       & 0.194                                & 0.238                            & 0.275                                  & 0.329                                 & 0.419                                  & 0.434                                 & 0.209                               & 0.263                               & 0.227                                 & 0.282                                 & 0.195                       & 0.247                                        & 0.406                               & 0.457                               & 0.225                               & 0.281                               & 0.219                                 & 0.261                                 \\
\multicolumn{1}{c|}{}                                                                & 336 & \textbf{0.231}                      & \textbf{0.258}                      & 0.237                           & 0.270                          & 0.245                                & 0.282                                  & 0.339                                  & 0.377                                 & 0.583                                  & 0.543                                 & 0.254                               & 0.301                               & 0.265                                 & 0.319                                 & 0.247                       & 0.306                                        & 0.462                               & 0.492                               & 0.278                               & 0.331                               & 0.280                                 & 0.306                                 \\
\multicolumn{1}{c|}{}                                                                & 720 & \textbf{0.305}                      & \textbf{0.312}                      & 0.311                                 & 0.327                                & 0.314                                & 0.325                            & 0.389                                  & 0.409                                 & 0.916                                  & 0.705                                 & 0.310                        & 0.341                               & 0.323                                 & 0.362                                 & 0.316                       & 0.351                                        & 0.517                               & 0.527                               & 0.311                               & 0.356                               & 0.365                                 & 0.359                                 \\ \hline
\end{tabular}
\end{adjustbox}
\label{main_result}
\end{table*}
When the model is used to make predictions over long time horizons the predictions starts to experience drift from the actual trajectory. In many use cases the measurement data is sparsely and irregularly available the long prediction window. Using a data assimilation-based framework we can make use of these measurements to correct the drift present in the predictions in a systematic way.

The extended Kalman filter (EKF) framework is a principled framework which updates the predicted states in a NLDS as a function of the error between the observed and predicted measurements~\cite{sarkka2023bayesian}.
Assuming data $Y_k$ is available, we can define the correction equation inspired by the EKF as
\begin{gather}
    \begin{bmatrix}
    Z'_k \\ R'_k
    \end{bmatrix}
    = \begin{bmatrix}
        \hat{Z}_k \\ R_k
    \end{bmatrix}
    + \begin{bmatrix}
        L_g & 0\\
        0 & L_r
    \end{bmatrix}
    \begin{bmatrix}
        J_{\psi}^T \\
        J_{\psi}^T
    \end{bmatrix}
    \begin{bmatrix}
        \tilde{Y}_k - Y_k
    \end{bmatrix} \,.
    \label{eq:correction_equation}
\end{gather}
Here $J_{\psi}$ is the Jacobian of the decoder function evaluated at the output of the prediction step i.e. $\hat{h}_{k} = \hat{Z}_k + R_k$. The Jacobian linearizes the nonlinear measurement function around the current predicted state:
\begin{equation}
    J_{\psi} = \frac{\partial \psi(\cdot;\theta)}{\partial h} \bigg|_{\hat{h}_{k}}
\end{equation}
The output of the correction step $h'_k=Z'_k + R'_k$ is then fed to the decoder to get the corrected prediction of the measurement trajectory.
\subsubsection{Kalman Gain}
The gain matrices $L_g$ and $L_r$ represent the component of Kalman gain for the global and residual branches. We use a gating mechanism parameterised by neural network to define the Kalman gains,
\begin{align}
    L_g &= \tanh(W_z^z(\hat{Z}_k) + W_y^g(\phi_g(Y^g_k))+ b_g)\,,
    \\
    L_r &= \tanh(W_z^r(R_k) + W_y^r(\phi_r(Y^r_k))+ b_r) \,.
    \label{eq:kalman_gain}
\end{align}
here $W_z^z, W_z^r, W_y^r, W_y^g$ are all single layer neural networks with ReLU activation. $b_r, b_g$ are bias terms for $L_r, L_g$ respectively.

\subsection{Learning criterion}

\subsubsection{Operator Learning Loss}
The physical model consists of the following trainable components: (i) latent Koopman model $\mathcal{K}$, (ii) state encoding model  and (iii) state decoding model. To ensure that the koopman dynamcis is respected by the learned physical model we minimise the following losses,
\begin{align*}
    \mathcal{L}_{recon} =\sum_{t=0}^H||\hat{y}_k - y_k||_2, 
    &\mathcal{L}_{pred} = \sum_{t=1}^T||\tilde{y}_k - y_k||_2,\\
    \mathcal{L}_{align} &= \sum_{k=1}^{T/k}||\hat{Z}_k - Z_k||_2
    \label{eq:loss}
\end{align*}
$H$ is forecast horizon and $T$ is lookback window length. To achieve the best performance we train KODA in stages. First we train the prediction model with $\mathcal{L}_{align}$, $\mathcal{L}_{pred}$ and $\mathcal{L}_{recon}$. We then add the correction model and train only using $\mathcal{L}_{pred}$. This two stage training process allows us to first learn the prediction model and then in the second stage learn the gain function while further improving the prediction output.
\subsubsection{Training Criteria for $L_g$ and $L_r$}
The Kalman gain $L_g$ and $L_r$ depends on whether the measurement data $Y_k$ is available for assimilation or not. The gating mechanism is first trained separately to allow correction to happen only when measurement data is made available during the inference time. During inference when no data is available the correction component should be $0$ such that \eqref{eq:kalman_gain} is only driven by the prediction term. For further details on training criteria and network architecture please refer to the appendix.

\section{Experimental setting}
\begin{table*}[!htbp]
\centering
\caption{Multivariate long-term time series forecasting with data assimilation results for KODA. We evaluate KODA across forecast horizon of \(H = 720\) in all datasets. We use $\alpha$ to denote $\%$ of measurement data available for online assimilation task in the forecast horizon. We report the mean of 10 runs for each value of $\alpha$.} 
\begin{adjustbox}{max width=.8\textwidth}
\begin{tabular}{c|llllllllllllll}
\hline
\multicolumn{1}{l|}{Alpha($\alpha$)} & \multicolumn{2}{c|}{ETTh1} & \multicolumn{2}{c|}{ETTh2} & \multicolumn{2}{c|}{ETTm1} & \multicolumn{2}{c|}{ETTm2} & \multicolumn{2}{c|}{Electricity} & \multicolumn{2}{c|}{Traffic} & \multicolumn{2}{c}{Weather} \\ \hline
               \%          & MSE          & MAE         & MSE          & MAE         & MSE          & MAE         & MSE          & MAE         & MSE             & MAE            & MSE           & MAE          & MSE          & MAE          \\ \hline
10                         & 0.397        & 0.439       & 0.323        & 0.402       & 0.391        & 0.420       & 0.311        & 0.358       & 0.181           & 0.293          & 0.411         & 0.268        & 0.367        & 0.295        \\
20                         & 0.378        & 0.402       & 0.318        & 0.489       & 0.355        & 0.413       & 0.289        & 0.341       & 0.175           & 0.278          & 0.352         & 0.245        & 0.336        & 0.277        \\
30                         & 0.351        & 0.370       & 0.296        & 0.362       & 0.340        & 0.392       & 0.268        & 0.305       & 0.173           & 0.271          & 0.328         & 0.219        & 0.319        & 0.258        \\ \hline
\end{tabular}
\end{adjustbox}
\label{assimilation_result}
\end{table*}
\begin{table}[!htbp]
\centering
\caption{State prediction for simple NLDS. KODA(G) represents the model with just the global Koopman component of KODA (without Fourier filtering). We compare different Koopman autoencoder-based models for thr state prediction task. These models are evaluated using the $100\times$ MSE over \(H=1000\).} 
\begin{adjustbox}{max width=.45\textwidth}
\begin{tabular}{l|cccc}
\hline
Data               & \multicolumn{1}{l}{KODA} & \multicolumn{1}{l}{KODA(G)} & \multicolumn{1}{l}{KAE} & \multicolumn{1}{l}{KAE (PE)}  \\ \hline
Pendulum           & 0.059                     & 0.106                        & 12.023                   & 0.214                                                     \\
Duffing Oscillator & 0.208                     & 0.451                        & 23.042                   & 1.198                                                    \\
Lotka-Volterra     & 0.119                     & 0.128                        & 2.956                    & 0.449                                                   \\
Lorenz' 63         & 32.288                    & 49.327                       & -                        & 83.416                                                 \\ \hline
\end{tabular}
\end{adjustbox}
\label{state_pred}
\end{table}

\begin{table}[ht]
\centering
\caption{We perform ablation study of the KODA framework. All results are for \(H=720\) on ETT benchmark. We compare four different variations of the KODA frameworks.}
\begin{adjustbox}{max width=0.47\textwidth}
\begin{tabular}{c|cccccccc}
\hline
Dataset                             & \multicolumn{2}{c}{ETTh1}                                               & \multicolumn{2}{c}{ETTh2}                                               & \multicolumn{2}{c}{ETTm1}                                               & \multicolumn{2}{c}{ETTm2}                                               \\ \cline{2-9} 
Metric                              & MSE                                & \multicolumn{1}{c|}{MAE}           & MSE                                & \multicolumn{1}{c|}{MAE}           & MSE                                & \multicolumn{1}{c|}{MAE}           & MSE                                & MAE                                \\ \hline
KODA(G (w. $L_g$))                      & 0.508                              & 0.536                              & 0.442                              & 0.479                              & 0.459                              & 0.468                              & 0.389                              & 0.418                              \\
KODA(R (w. $L_r$))                      & 0.451                              & 0.497                              & 0.406                              & 0.436                              & 0.453                              & 0.455                              & 0.395                              & 0.422                              \\
KODA(P w/o C)                       & 0.429                              & 0.458                              & 0.379                              & 0.425                              & 0.414                              & 0.423                              & 0.329                              & 0.360                              \\
KODA(w/o L)                         & 0.417                              & 0.454                              & \textbf{0.358}                              & 0.418                              & 0.421                              & 0.426                              & 0.328                              & 0.362                              \\ \hline
\multicolumn{1}{l|}{KODA(complete)} & \multicolumn{1}{l}{\textbf{0.411}} & \multicolumn{1}{l}{\textbf{0.451}} & \multicolumn{1}{l}{0.362} & \multicolumn{1}{l}{\textbf{0.416}} & \multicolumn{1}{l}{\textbf{0.409}} & \multicolumn{1}{l}{\textbf{0.420}} & \multicolumn{1}{l}{\textbf{0.322}} & \multicolumn{1}{l}{\textbf{0.358}} \\ \hline
\end{tabular}
\end{adjustbox}
\label{abation}
\end{table}
We evaluate KODA across multiple datasets and conduct an extensive comparison across different baselines. We evaluate KODA over two main tasks: \textbf{(1)} Forecasting and \textbf{(2)} Data Assimilation. We also present comparison based on some well-known NLDS datasets for the \textbf{(3)} State Prediction task.  Through our experiments we showcase the comparable effectiveness of KODA with state of the art baselines for forecasting and then further improvement with the data assimilation strategy.

\subsubsection{Datasets:}
We compare KODA across 4 different real world multivariate NLDS time series datasets : \textbf{(1)} weather (Wetterstation), \textbf{(2)} traffic (PeMS), \textbf{(3)} ECL (UCI), \textbf{(4)} ETT \cite{zhou2021informer}(including 4 subsets: ETTh1, ETTh2, ETTm1, ETTm2). We also evaluate the performance of KODA on M4 \cite{SpyrosMakridakis} which is a real world univariate time series on periodically collected marketing data. We use the same data prepossessing and test time split ratio as presented in \cite{wutimesnet}. 

We use the same datasets for showing results for data assimilation task. We present additional results on state prediction task for some of the well known uni-variate NLDS: \textbf{(1)} Lokta Volterra, \textbf{(2)} Duffing Oscillator, \textbf{(3)} pendulum and \textbf{(4)} Lorenz 63. These NLDS present extremely nonlinear dynamics with varying dimensionality including multiple fixed points. Appendix B for NLDS description. 
\subsubsection{Baseline:}
We choose state of the art methods to compare KODA  performance on forecasting and data assimilation tasks. We would like to make the note that most deep learning methods discussed as our baseline don't have a well defined way for data assimilation. For the comparison on the forecasting task we evaluate our model against MLP based methods: Koopman forecaster (KNF)\cite{wang2023koopman}, Koopa \cite{liu2024koopa}, Dlinear \cite{Zeng2022AreTE}, TiDE \cite{das2023longterm}; Transformer based methods: fedformer \cite{zhou2022fedformer}, PatchTST \cite{nietime}; Informer \cite{zhou2021informer} Temporal convolution method: TimesNet \cite{wutimesnet}, MICN \cite{wang2023micn}. For the state prediction task from noisy measurements on the known NLDS we compare KODA against vanilla Koopman autoencoder \cite{lusch2018deep}, MLP and Koopman autoencoder with periodic re-encoding \cite{fathi2024course}. We compare KODA against the method presented in \cite{frion2024neural} for the data assimilation and forecasting task when data is sparsely present in the lookback window. KODA's framework allows us to assimilate measurement data into prediction recursively. Hence, to evaluate KODA's performance we use different amount of noisy measurement present in the forecast window and evaluate model forecast on the rest of the forecast window. We compare this over different sets of noise levels in the measurement data.

\subsubsection{Evaluation Metrics:} 
The most commonly used metric for comparing the accuracy of the forecasting method is: \textbf{(1)} Mean Absolute Error (\textbf{MAE}) and \textbf{(2)} Mean Squared Error (\textbf{MSE}).
For further details on the experimental setup, please refer to Appendix B.

\section{Results}
\subsubsection{Long Term Forecast}
In Table \ref{main_result} we present the results for long term forecasting across several recent state-of-the-art methods. We compare all the methods on four different multivariate time series datasets. The choice of horizon $T_H$ is set to be $\{96,192,336,720\}$. Through this task we are solely evaluating KODA's ability to generate long term accurate forecasts. The lookback window for all the models in Table \ref{main_result} was set to be $T_L=720$. We present additional results across different lookback windows in the appendix. We observe that except for the traffic benchmark, KODA almost beats all the other methods across benchmarks. In the case of traffic dataset we observe that the best performing model were TiDE and PatchTST across several of the prediction horizon length except when $H = 720$, where KODA provides the second best performance. One interesting observation is that KODA outperforms all the other Koopman-based models: KOOPA and KNF. This shows that KODA presents a flexible model that can produce long term accurate forecast for multivariate non-stationary time series datasets. We provide additional results including results on uni-variate time-series benchmark M4 \cite{SpyrosMakridakis} in Appendix C.
\subsubsection{Data Assimilation}
One of the main contribution of KODA is in its ability to perform online data assimilation during inference time hence making the model highly adaptive. In Table \ref{assimilation_result} we present the result for the online data assimilation task. We define $\alpha$ as the percentage of data available in the forecast horizon for assimilation. We show results for $\alpha \in \{10\%, 20\%, 30\% \}$ across all the benchmarks. For this experiment we chose $\tau=24$ and $w = 8$. For the sake of simplicity we distributed the available measurements along the prediction horizon uniformly. We see a consistent improvement in the forecast results when correction is used. This shows that KODA other than being able to generate accurate long term forecasts, KODA is also able to make use of the data available at a future time to further improve its forecast. This makes KODA framework really potent in tasks where data is irregularly present at different intervals and long term forecasting is required. Additionally KODA can also be used when the data are irregularly present instead of uniformly. We show additional results on assimilation task across benchmarks in the appendix C.

\subsubsection{State Prediction}
The main attraction of using Koopman operators based approaches is achieving system identification in a complete data driven approach. To demonstrate KODA's ability for such, we generate 100 trajectories of the different NLDS mentioned above. For further details on data generation process and NLDS desciption refer appendix A. We train the model using first 500 steps of the generated trajectories. During inference we generate the forecast for 1000 steps from the model. We observe that the model is able to generalise to the part of state space that were previously not seen during the training. We present the results in Table \ref{state_pred}. Here we compare KODA with Koopman autoencoder methods proposed in literature \cite{lusch2018deep,fathi2024course}. We also evaluate KODA and KODA (G) separately to evaluate the effectiveness of the physical branch of KODA. We observe that not only KODA but KODA (G) performs much better than KAE and KAE (PE). This attributed to the structure of our training procedure that doesn't enforce the whole trajectory in the latent space to be globally linear but only enforces local linearity in the latent space. This is similar to the course correction achieved in \cite{fathi2024course} using periodic re-encoding. Yet our model is largely different from the one used in \cite{fathi2024course}. We can observe that both KODA and KODA (G) performs better than KAE(PE) on all four datasets. 

\subsubsection{Ablation Study}
To compare the efficacy of different elements of KODA we evaluate 5 different variations of KODA framework across different subsets of ETT dataset. These variations include \textbf{1.} global model of KODA with the: KODA (G (w. $L_g$)); \textbf{2.} Residual model of KODA: KODA (R (w. $L_r$)); \textbf{3.} Only the prediction model without any correction component: KODA (P w/o C); \textbf{4.} Full KODA model but without the Kalman gain gating function $L$: KODA (w/o $L$) and finally \textbf{5.} Full KODA model. For KODA (G (w. $L_g$)) and KODA (R (w. $L_r$)) we also don't use the measurement disentanglement. We present the results of the ablation study in Table \ref{abation}. When we only use the residual component, KODA reduces to an implementation of SegRNN \cite{lin2023segrnn}. Albeit in \cite{lin2023segrnn} there is no data assimilation framework. KODA(complete) gives the best performance. We also observe a significant jump when the whole prediction model is used versus individual branches. We provide further ablation study of the KODA framework in appendix C.

\section{Discussion}
In this paper we proposed a novel framework for time series forecasting and data assimilation named KODA. Through rigorous experimentation across different benchmarks we showed that KODA is able to \textbf{(1)} Generate accurate long term forecast, \textbf{(2)} Generate a disentangled view of the NLDS into local and global components, \textbf{(3)} Use a correction framework to perform online data assimilation, \textbf{(4)} Perform state prediction of the state space for known NLDS. Therefore KODA present a flexible recursive framework that is able to exploit the benefits of koopman operator representation while being able to generate long term forecast and perform data assimilation.
\appendix

\bibliography{aaai25}

\end{document}